\documentclass[runningheads]{llncs}
\usepackage{graphicx}
\usepackage[T1]{fontenc}
\usepackage{graphicx}
\usepackage{graphicx}
\usepackage{algorithmic}
\usepackage{amssymb}
\usepackage{algorithm}
\usepackage{color}
\usepackage{colortbl}
\usepackage{siunitx}
\usepackage{booktabs}
\usepackage{threeparttable}
\usepackage{multirow}
\usepackage{marvosym}
\usepackage{amsmath}
\usepackage[colorlinks,hyperindex,breaklinks]{hyperref}
\usepackage{amssymb} % http://ctan.org/pkg/amssymb
\usepackage{pifont} % http://ctan.org/pkg/pifont
\usepackage{arydshln}
\definecolor{anti-flashwhite}{rgb}{0.95, 0.95, 0.96}
\definecolor{whitesmoke}{rgb}{0.94, 0.94, 0.94}
\definecolor{teagreen}{rgb}{0.82, 0.94, 0.75}
\definecolor{powderblue}{rgb}{0.69, 0.88, 0.9}
\definecolor{pastelblue}{rgb}{0.68, 0.78, 0.81}
\definecolor{lightskyblue}{rgb}{0.53, 0.81, 0.98}
\definecolor{turquoise}{cmyk}{0.65,0,0.1,0.3}
\definecolor{purple}{rgb}{0.65,0,0.65}
\definecolor{dark_green}{rgb}{0, 0.5, 0}
\definecolor{orange}{rgb}{0.8, 0.6, 0.2}
\definecolor{red}{rgb}{0.8, 0.2, 0.2}
\definecolor{darkred}{rgb}{0.6, 0.1, 0.05}
\definecolor{blueish}{rgb}{0.0, 0.3, .6}
\definecolor{light_gray}{rgb}{0.7, 0.7, .7}
\definecolor{pink}{rgb}{1, 0, 1}
\definecolor{greyblue}{rgb}{0.25, 0.25, 1}

\newcommand{\ie}{\textit{i}.\textit{e}.}
\newcommand{\eg}{\textit{e}.\textit{g}.}
\newcommand{\etal}{\textit{et al}.}

\makeatletter

\newcommand{\Rmnum}[1]{\expandafter\@slowromancap\romannumeral #1@}
\makeatother

\begin{document}
% ----------------------------------------
\title{Few Shot Medical Image Segmentation with Cross Attention Transformer}
% ----------------------------------------
% camera-ready
% ----------------------------------------
\author{Yi Lin$^{\dag}$, 
%index{Last Name, First Name}
Yufan Chen$^{\dag}$, 
%index{Last Name, First Name}
Kwang-Ting Cheng, 
%index{Last Name, First Name}
and Hao Chen\textsuperscript{\Letter}}
\institute{The Hong Kong University of Science and Technology, Hong Kong, China \\
\email{jhc@cse.ust.hk}}
\titlerunning{Few Shot Medical Image Segmentation with Cross Attention Transformer}
\authorrunning{Y. Lin,~Y. Chen~\etal}
\def\thefootnote{$\dag$}\footnotetext{Equal contribution; \Letter~corresponding author.}
\maketitle % typeset the header of the contribution
%------------------------------------------
% abstract
\begin{abstract}
Medical image segmentation has made significant progress in recent years. Deep learning-based methods are recognized as data-hungry techniques, requiring large amounts of data with manual annotations. 
However, manual annotation is expensive in the field of medical image analysis, which requires domain-specific expertise. 
To address this challenge, few-shot learning has the potential to learn new classes from only a few examples.
In this work, we propose a novel framework for few-shot medical image segmentation, termed CAT-Net, based on cross masked attention Transformer. 
Our proposed network mines the correlations between the support image and query image, limiting them to focus only on useful foreground information and boosting the representation capacity of both the support prototype and query features. 
We further design an iterative refinement framework that refines the query image segmentation iteratively and promotes the support feature in turn.
We validated the proposed method on three public datasets: Abd-CT, Abd-MRI, and Card-MRI. 
Experimental results demonstrate the superior performance of our method compared to state-of-the-art methods and the effectiveness of each component.
Code: \url{https://github.com/hust-linyi/CAT-Net}.
\end{abstract}
%------------------------------------------
% introduction
\section{Introduction}
\label{sec:introduction}
Automatic segmentation of medical images is a fundamental step for a variety of medical image analysis tasks, such as diagnosis, treatment planning, and disease monitoring~\cite{che2023DGDR,che2023iqad}.
The emergence of deep learning (DL) has enabled the development of many medical image segmentation methods, which have achieved remarkable success~\cite{transunet,lin2023rethinking,zhang2022deep,che2022learning,lin2021seg4reg+}.
 Most of the existing methods follow a fully-supervised learning paradigm, which requires a considerable amount of labeled data for training. 
However, the manual annotation of medical images is time-consuming and labor-intensive, limiting the application of DL in medical image segmentation. 
Specifically for the 3D volumetric medical images (\eg, CT, MRI), the manual annotation is even more challenging which requires the annotators to go through hundreds of 2D slices for each 3D scan.

To address the challenge of manual annotation, various label-efficient techniques have been explored, such as self-supervised learning~\cite{ouyang2022self}, semi-supervised learning~\cite{xu2022all,yang2019bi}, and weakly-supervised learning~\cite{lin2022label}. 
Despite leveraging information from unlabeled or weakly-labeled data, these techniques still require a substantial amount of training data~\cite{17siam2019amp,pan2023human}, which may not be practical for novel classes with limited examples in the medical domain.
This limitation encourages the few-shot learning paradigm~\cite{8snell2017prototypical,9sung2018learning,10garcia2017few,11vinyals2016matching} to be applied to medical image segmentation.
Specifically, the few-shot learning paradigm aims to learn a model from a small number of labeled data (denoted as \textit{support}) and then apply it to a new task (denoted as \textit{query}) with only a few labeled data without any retraining.
Considering the hundreds of organs and countless diseases in the human body, FSL brings great potential to the various medical image segmentation tasks where a new task can be easily investigated in a data-efficient manner.

Most few-shot segmentation methods follow the learning-to-learn paradigm, which aims to learn a meta-learner to predict the segmentation of query images based on the knowledge of support images and their respective segmentation labels.
The success of this paradigm depends on how effectively the knowledge can be transferred from the support prototype to the query images.
Existing few-shot segmentation methods mainly focus on the following two aspects: (1) how to learn the meta-learner~\cite{pandey2022adversarially,tian2022gfsseg,pfenetpp}; and (2) how to better transfer the knowledge from the support images to the query images~\cite{SUN2022105067,tian2020pfenet,peng2023hierarchical,protomineing,self-support,Tang_2021_ICCV}.
Despite prototype-based methods having shown success, they typically ignore the interaction between support and query features during training.
In this paper, as shown in Fig.~\ref{fig:framework}(a), we propose \textbf{CAT}-\textbf{Net}, a \textbf{C}ross \textbf{A}ttention \textbf{T}ransformer network for few-shot medical image segmentation, which aims to fully capture intrinsic classes details while eliminating useless pixel information and learn an interdependence between the support and query features.
Different from the existing FSS methods that only focus on the single direction of knowledge transfer (\ie, from the support features to the query features), the proposed CAT-Net can boost the mutual interactions between the support and query features, benefiting the segmentation performance of both the support and query images. 
Additionally, we propose an iterative training framework that feed the prior query segmentation into the attention transformer to effectively enhance and refine the features as well as the segmentation.
Three publicly available datasets are adopted to evaluate our CAT-Net, \ie, Abd-CT~\cite{landman2015miccai}, Abd-MRI~\cite{kavur2021chaos}, and Card-CT~\cite{zhuang2018multivariate}. 
Extensive experiments validate the effectiveness of each component in our CAT-Net, and demonstrate its state-of-the-art performance.
%------------------------------------------
% method
\section{Method}
\subsection{Problem Definition}
Few-shot segmentation (FSS) aims to segment novel classes by just a few samples with densely-annotated samples. 
In FSS, the dataset is divided into the training set $\mathbb{D}_\text{train}$, containing the base classes $\mathbb{C}_\text{train}$, and the test set $\mathbb{D}_\text{test}$, containing the novel classes $\mathbb{C}_\text{test}$, where $\mathbb{C}_\text{train}\cap \mathbb{C}_\text{test}= \emptyset$.
To obtain the segmentation model for FSS, the commonly used episode training approach is employed~\cite{wang2019panet}.
Each trainig/testing episode ($S_i, Q_i$) instantiates a $N$-way $K$-shot segmentation learning task.
Specifically, the support set $S_i$ contains $K$ samples of $N$ classes, while the query set $Q_i$ contains one sample from the same class.
The FSS model is trained with episodes to predict the novel class for the query image, guided by the support set.
During inference, the model is evaluated directly on $\mathbb{D}_\text{test}$ without any re-training.
In this paper, we follow the established practice in medical FSS~\cite{hansen2022anomaly,ouyang2022self,shen2022q} that consider the \textbf{1}-way \textbf{1}-shot task.

\begin{figure}[!t]
\centering\includegraphics[width=\textwidth]{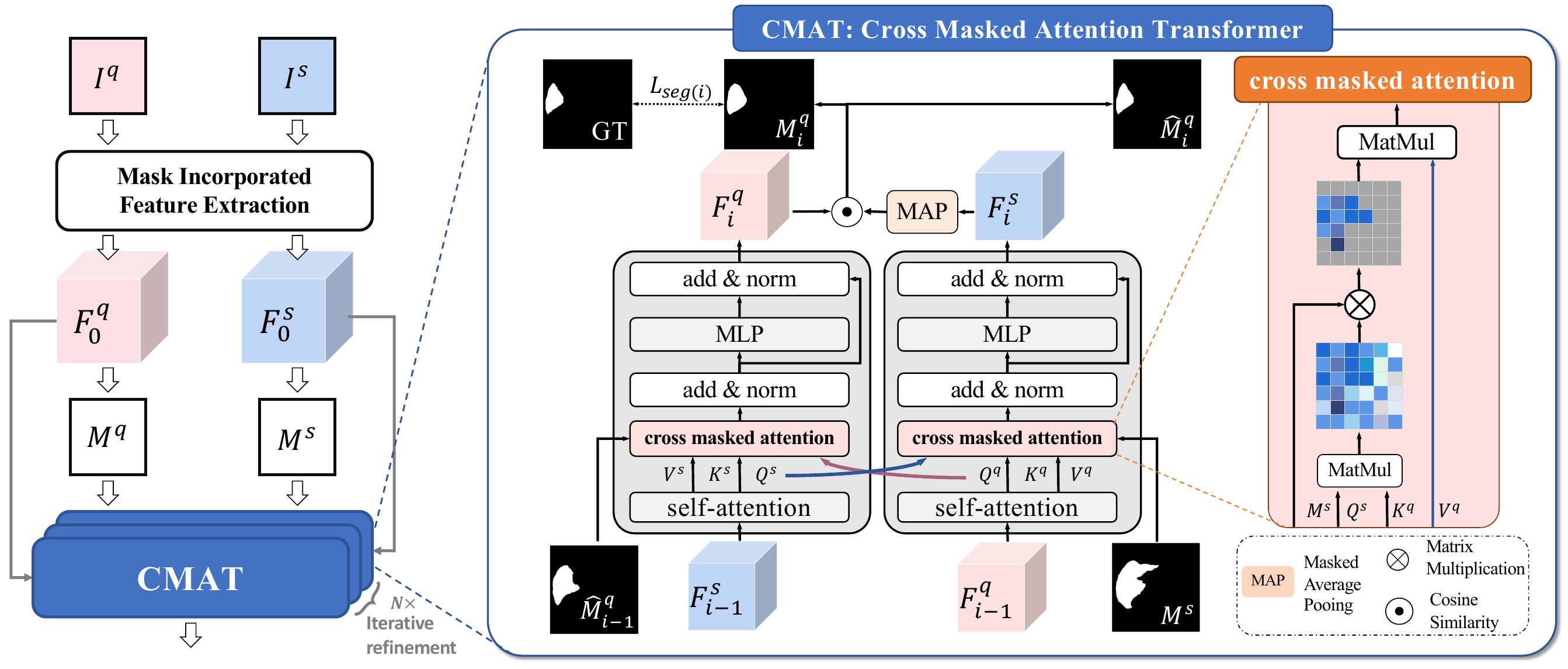}
\caption{(a) Overview of the CAT-NET; (b) The architecture of CMAT module.}
\label{fig:framework}
\end{figure}
\subsection{Network Overview}
The Overview of our CAT-Net is illustrated in Fig.~\ref{fig:framework}(a). It consists of three main components:
1) a mask incorporated feature extraction (MIFE) sub-net that extracts initial query and support features as well as query mask; 
2) a cross masked attention Transformer (CMAT) module in which the query and support features boost each other and thus refined the query prediction;
and 3) an iterative refinement framework that sequentially applies the CMAT modules to continually promote the segmentation performance.
The whole framework can be trained in an end-to-end fashion.

\subsection{Mask Incorporated Feature Extraction}
The Mask Incorporate Feature Extraction (MIFE) sub-net takes query and support images as input and generates their respective features, integrated with the support mask. A simple classifier is then used to predict the segmentation for the query image.
Specifically, we first employ a feature extractor network (\ie, ResNet-50) to map the query and support image pair $I^q$ and $I^s$ into the feature space, producing multi-level feature maps $F^q$ and $F^s$ for query and support image, respectively.
Next, the support mask is pooled with $F^s$ and then expanded and concatenated with both $F^q$ and $F^s$.
Additionally, the segmentation mask of query image in MIFE is further concatenated with the query feature to strengthen the correlation between query and support features via a pixel-wise similarly map. 
Finally, the query feature is processed by a simple classifier to get the query mask.
Further details of the MIFE architecture can be found in the supplementary material.

\subsection{Cross Masked Attention Transformer}
As shown in Fig.~\ref{fig:framework}(b), the cross masked attention Transformer (CMAT) module comprises three main components: 1) a self-attention module for extracting global information from query and support features; 2)  a cross masked attention module for transferring foreground information between query and support features while eliminating redundant background information, and 3) a prototypical segmentation module for generating the final prediction of the query image.

\noindent\textbf{Self-Attention Module.}
To capture the global context information of every pixel in the query feature $F_0^q$ and support features $F_0^s$, the initial features are first flattened into 1D sequences and fed into two identical self-attention modules.
Each self-attention module consists of a multi-head attention (MHA) layer and a multi-perceptron (MLP) layer. 
Given an input sequence $S$, the MHA layer first projects the sequence into three sequences $K$, $Q$, and $V$ with different weights. The attention matrix $A$ is then calculated as:
\begin{equation}
A(Q, K)=\dfrac{QK^T}{\sqrt{d}}
\end{equation}
where $d$ is the dimension of the input sequence.
The attention matrix is then normalized by a softmax function and multiplied by the value sequence $V$ to get the output sequence $O$:
\begin{equation}
O=\text{softmax}(A)V
\end{equation}
The MLP layer is a simple $1\times 1$ convolution layer that maps the output sequence $O$ to the same dimension as the input sequence $S$.
Finally, the output sequence $O$ is added to the input sequence $S$ and normalized using layer normalization (LN) to obtain the final output sequence $X$.
The output feature sequence of the self-attention alignment encoder is represented by $X^q\in\mathbb{R}^{HW\times D}$ and $X^s\in\mathbb{R}^{HW\times D}$ for query and support features, respectively.

\noindent\textbf{Cross Masked Attention Module.}
We utilize cross masked attention to incorporate query features and support features with respect to their foreground information by constraining the attention region in attention matrix with support and query masks. 
Specifically, given the query feature $X^q$ and support features $X^s$ from the aforementioned self-attention module, we first project the input sequence into three sequences $K$, $Q$, and $V$ using different weights, resulting in $K^q$, $Q^q$, $V^q$, and $K^s$, $Q^s$, $V^s$, respectively. 
Taking the support features as an example, the cross attention matrix is calculated by:
\begin{equation}
\text{A}(K^q,Q^s)=\dfrac{(K^q)^T Q^s}{\sqrt{d}}
\end{equation}
We expand and flatten the binary query mask $M^q$ to limit the foreground region in attention map. 
The masked cross attention (MCA) map is computed as:
\begin{equation}
\text{MCA}(K^q,Q^s,V^q, {M}^s)=M^s \cdot V^q(\text{softmax}(A(K^q,Q^s)))
\end{equation}
Similar to self-attention, the support feature is processed by MLP and LN layer to get the final enhanced query features $F^s_1$.
Similarly, the enhanced query feature $F^q_1$ is obtained with foreground information from the query feature.

\noindent\textbf{Prototypical Segmentation Module.}
Once the enhanced query and support features are obtained, the prototypical segmentation is used to obtain the final prediction.
First, a prototype of class $c$ is built by masked average pooling of the support feature $F^s_1$ as follows:
\begin{equation}
p_c = \dfrac{1}{K}\sum^K_{k=1}\dfrac{\sum_{k,x,y} F^s_{i,(k,x,y)}m^s_{(k,x,y,c)}}{\sum_{x,y}m^s_{(k,x,y,c)} }
\end{equation}
where $K$ is the number of support images, and $m^s_{(k,x,y,c)}$ is a binary mask that indicates whether pixel at the location $(x,y)$ in support feature $k$ belongs to class $c$.
Next, we use the non-parametirc metric learning method to perform segmentation. 
The prototype network calculates the distance between the query feature vector and the prototype $P={\{P_c|c\in C\}}$. 
Softmax function is applied to produce probabilistic outputs for all classes, generating the query segmentation:
\begin{equation}
\hat{M}^q_{i,(x,y)} = \text{softmax}\big(\alpha \text{cos}(F^q_{i,(x,y)},p_c)\cdot{\text{softmax}(\alpha \text{cos}(F^q_{i,(x,y)},p_c))}\big)
\end{equation}
where $\text{cos}(\cdot)$ denotes cosine distance, $\alpha$ is a scaling factor that helps gradients to back-propagate in training. In our work, $\alpha$ is set to 20, same as in~\cite{wang2019panet}.

Additionally, we design a double threshold strategy to obtain query segmentation.
Specifically, we set the first threshold $\tau$ to 0.5 to obtain the binary query mask ${M}^q$, which is used to calculate the Dice loss and update the model. 
Then, the second threshold $\hat{\tau}$ is set to 0.4 to obtain the dilated query mask $\hat{M}^q$, which is used to generate the enhanced query feature $F^q_2$ in the next iteration.
The second threshold $\hat{\tau}$ is set lower than the first threshold $\tau$ to prevent some foreground pixels from being mistakenly discarded.
The query segmentation mask ${M}^q$ and dilated mask $\hat{M}^q$ are represented by:
\begin{equation}
{M}^q_i=
\left\{
\begin{aligned}
1, \quad &M^q_{i,(x,y)} > \tau\\
0,\quad & M^q_{i,(x,y)} < \tau \\
\end{aligned}
\right.
\qquad
\hat{M}^q_i=
\left\{
\begin{aligned}
1, \quad &M^q_{i,(x,y)} > \hat{\tau}\\
0,\quad & M^q_{i,(x,y)} < \hat{\tau} \\
\end{aligned}
\right.
\end{equation}

\subsection{Iterative Refinement framework}
As explained above, the CMAT module is designed to refine the query and support features, as well as the query segmentation mask. 
Thus, it's natural to iteratively apply this sub-net to get the enhanced features and refine the mask, resulting in a boosted segmentation result. 
The result after the $i$-th iteration is represented by:
\begin{equation}
(F^s_i,F^q_i,M^q_i,\hat{M}^q_i)=\text{CMAT}(F_{i-1}^{s}, F_{i-1}^{q},\hat{M}_{i-1}^{q},M^s)
\end{equation}
The subdivision of each step can be specifically expressed as:
\begin{equation}
(F^s_i,F^q_i)=\text{CMA}(F_{i-1}^{s}, F_{i-1}^{q},\hat{M}_{i-1}^{q},M^s)
\end{equation}
\begin{equation}
(M^q_i,\hat{M}_{i}^{q}) = \text{Proto}(F_{i}^{s}, F_{i}^{q},M^s, \tau,\hat{\tau})
\end{equation}
where $\text{CMA}(\cdot)$ indicates the self-attention and cross masked attention module, and $\text{Proto}(\cdot)$ represents the prototypical segmentation module.
%------------------------------------------
% experiment
\section{Experiment}
\label{sec:experiment}
\subsection{Dataset and Evaluation Metrics}
We evaluate the proposed method on three public datasets, \ie, Abd-CT~\cite{landman2015miccai}, Abd-MRI~\cite{kavur2021chaos}, and Card-MRI~\cite{zhuang2018multivariate}. 
Abd-CT contains 30 abdominal CT scans with annotations of left and right kidney (LK and RK), spleen (Spl), liver (Liv).
Abd-MRI contains 20 abdominal MRI scans with annotations of the same organs as Abd-CT.
Card-MRI includes 35 cardiac MRI scans with annotations of left ventricular blood pool (LV-B), left ventricular myocardium (LV-M), and right ventricle (RV).
We use the Dice score as the evaluation metric following~\cite{ouyang2022self,shen2022q}.

To ensure a fair comparison, all the experiments are conducted under the 1-way 1-shot scenario using 5-fold cross-validation. 
We follow~\cite{ouyang2022self} to remove all slices containing test classes during training to ensure that the test classes are all unseen during validation.
In each fold, we follow~\cite{ouyang2022self,hansen2022anomaly,shen2022q} that takes the last patient as the support image and the remaining patients as the query (setting \Rmnum{1}).
We further propose a new validation setting (setting \Rmnum{2}) that takes every image in each fold as a support image alternately and the other images as the query.
The averaged result of each fold is reported. 
It could evaluate the generalization ability of the model by reducing the affect of support image selection.

\subsection{Implementation Details}
The proposed method is implemented using PyTorch.
Each 3D scan is sliced into 2D slices and reshaped into 256$\times$256 pixels. 
Common 3D image pre-processing techniques, such as intensity normalization and resampling, are applied to the training data.
We apply episode training with 20$k$ iterations. SGD optimizer is adopted with a learning rate of 0.001 and a batch size of 1.
Each episode training takes approximately 4 hours using a single NVIDIA RTX 3090 GPU.

\begin{table}[thbp]
\centering
\small
\caption{Comparison with state-of-the-art methods in Dice coefficient (\%) on Abd-CT and Abd-MRI, and Card-MRI datasets under setting \Rmnum{1} \& \Rmnum{2}.}
\renewcommand\arraystretch{1.2}
\resizebox{\textwidth}{!}{
\begin{tabular}{c|c | c c c c c | c c c c c | c c c c } 
\multicolumn{1}{c}{} &  \multicolumn{1}{c}{} & \multicolumn{5}{c}{Abd-CT~\cite{kavur2021chaos}} & \multicolumn{5}{c}{Abd-MRI~\cite{landman2015miccai}} & \multicolumn{4}{c}{Card-MRI~\cite{zhuang2018multivariate}} \\
\toprule
\multicolumn{1}{c}{} & Methods & LK & RK & Spl. & Liv. & Avg. & LK & RK & Spl. & Liv. & Avg. & LV-B & LV-M & RV & Avg. \\
\hline 
\multirow{6}{*}{\rotatebox{90}{Setting \Rmnum{1}}} & SE-Net~\cite{roy2020squeeze} & 32.83 & 14.84 & 0.23 & 0.27 & 11.91 &62.11 & 61.32 & 51.80 & 27.43 & 50.66 & 58.04 & 25.18 & 12.86 & 32.03\\
& PA-Net~\cite{wang2019panet} & 37.58 & 34.69 & 43.73 & 61.71 & 44.42 & 47.71 & 47.95 & 58.73 & 64.99 & 54.85 & \textbf{70.43} & 46.79 & 69.52 & 62.25\\
& ALP-Net~\cite{ouyang2022self} & 63.34 & 54.82 & 60.25 & 73.65 & 63.02 & 73.63 & \underline{78.39} & 67.02 & 73.05 & 73.02 & 61.89 & 87.54 & 76.71 & 75.38\\
& AD-Net~\cite{hansen2022anomaly} & \textbf{63.84} & 56.98 & 61.84 & 73.95& 64.15 & 71.89 & 76.02 & 65.84 & 76.03 & 72.70 & 65.47 & 88.36 & 78.35 & 77.39\\
& Q-Net~\cite{shen2022q} & 63.26	& \underline{58.37}	& \underline{63.36} & \underline{74.36} &  \underline{64.83} & \textbf{74.05} & 77.52 & \underline{67.43} & \underline{78.71} & \underline{74.43} & \underline{66.87} & \underline{89.63} & \underline{79.25} & \underline{78.58}\\
 & \cellcolor{whitesmoke}\textbf{Ours} & \cellcolor{whitesmoke}\underline{63.36} & \cellcolor{whitesmoke}\textbf{60.05} & \cellcolor{whitesmoke}\textbf{67.65}	& \cellcolor{whitesmoke}\textbf{75.31}	& \cellcolor{whitesmoke}\textbf{66.59} & \cellcolor{whitesmoke}\underline{74.01} & \cellcolor{whitesmoke}\textbf{78.90}& \cellcolor{whitesmoke}\textbf{68.83} & \cellcolor{whitesmoke}\textbf{78.98} & \cellcolor{whitesmoke}\textbf{75.18} & \cellcolor{whitesmoke}66.85 & \cellcolor{whitesmoke}\textbf{90.54} & \cellcolor{whitesmoke}\textbf{79.71} & \cellcolor{whitesmoke}\textbf{79.03}\\
 \hline
\multirow{4}{*}{\rotatebox{90}{Setting \Rmnum{2}}} 
& ALP-Net~\cite{ouyang2022self} & 65.99 & 59.49 & 65.02 & 73.50 & 66.05 & 70.17 & 77.05 & \underline{67.71} & 72.45 & 71.85 & 61.61 & 87.13 & 77.35 & 75.36\\
& AD-Net~\cite{hansen2022anomaly} & \underline{67.35} & 59.88 & 64.35 & 76.78 & 67.09 & 72.26 & 76.57 & \textbf{67.89} & \underline{73.96} & 72.67 & 65.08 & 86.26 & 76.50 & 75.95\\
& Q-Net~\cite{shen2022q} & 66.25 & \underline{62.36} &\textbf{67.35}  & \underline{77.33} & \underline{68.32} & \underline{73.96} & \underline{81.07} & 65.39 & 72.36 & \underline{73.20} & \underline{66.35} & \underline{88.40} & \underline{79.37} & \underline{78.04}\\
& \cellcolor{whitesmoke}\textbf{Ours}& \cellcolor{whitesmoke}\textbf{68.82} & \cellcolor{whitesmoke}\textbf{64.56} & \cellcolor{whitesmoke}\underline{66.02} & \cellcolor{whitesmoke}\textbf{80.51} & \cellcolor{whitesmoke}\textbf{70.88} & \cellcolor{whitesmoke}\textbf{75.31} & \cellcolor{whitesmoke}\textbf{83.23} & \cellcolor{whitesmoke}67.31 & \cellcolor{whitesmoke}\textbf{75.02} & \cellcolor{whitesmoke}\textbf{75.22} & \cellcolor{whitesmoke}\textbf{67.21} & \cellcolor{whitesmoke}\textbf{90.54} & \cellcolor{whitesmoke}\textbf{80.34}  & \cellcolor{whitesmoke}\textbf{79.36} \\
\bottomrule
\end{tabular}
\label{tab:sota}}
\end{table}
\subsection{Comparison with State-of-the-Art Methods}
\begin{figure}[thbp]
\centering\includegraphics[width=\textwidth]{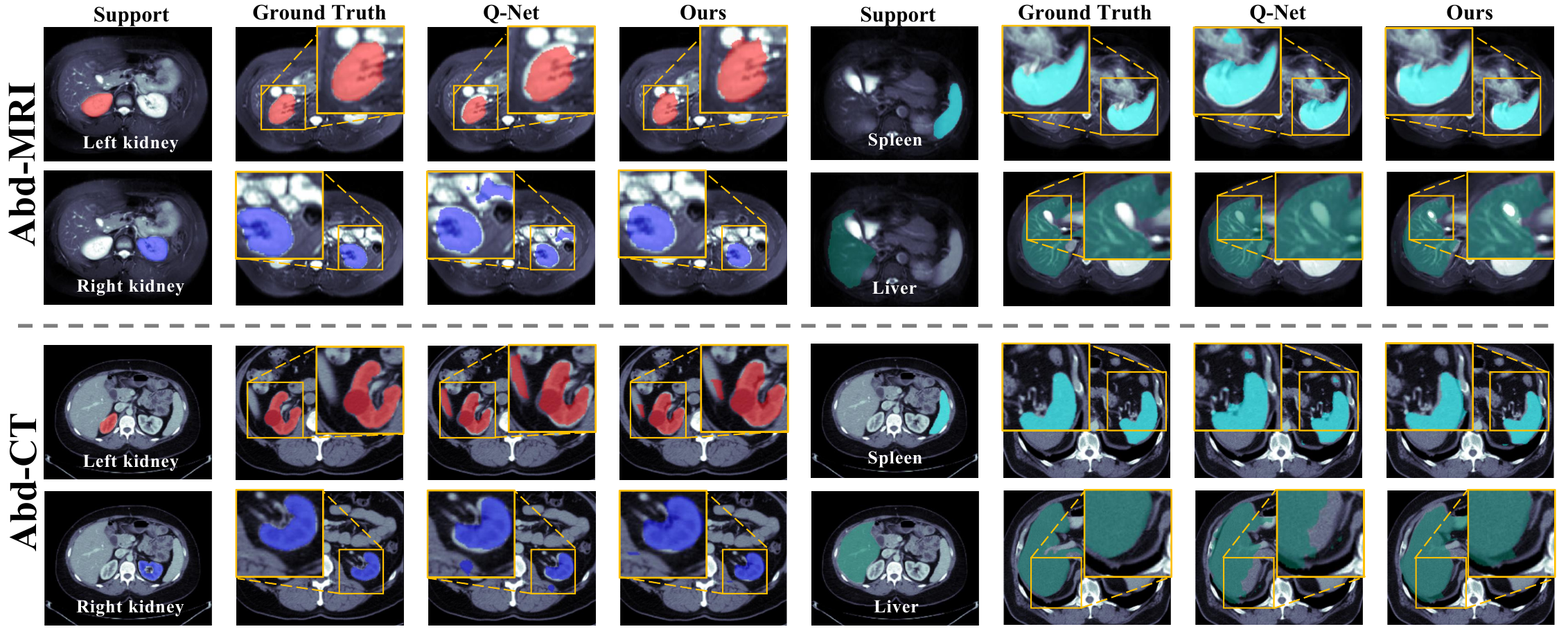}
\caption{Qualitative results of our method on Abd-CT and Abd-MRI.}
\label{fig:vision}
\end{figure}

We compare the proposed CAT-Net with state-of-the-art (SOTA) methods, including SE-Net~\cite{roy2020squeeze}, PANet~\cite{wang2019panet}, ALP-Net~\cite{ouyang2022self}, and AD-Net~\cite{hansen2022anomaly}, and Q-Net~\cite{shen2022q}. 
PANet~\cite{wang2019panet} are the typical prototypical FSS method in the natural image domain, SE-Net~\cite{roy2020squeeze}, ALP-Net~\cite{ouyang2022self}, AD-Net~\cite{hansen2022anomaly}, and Q-Net~\cite{shen2022q} are the most representative work in medical FSS task.
Experiment results presented in Table~\ref{tab:sota} demonstrate that the proposed method outperforms SOTAs on all three datasets under both setting \Rmnum{1} and setting \Rmnum{2}.
Under setting \Rmnum{1}, the proposed CAT-Net achieves 66.59\% Dice on Abd-CT, 75.18\% Dice on Abd-MRI, and 79.03\% Dice on Card-MRI in Dice, outperforming SOTAs by 1.76\%, 0.75\%, and 0.45\%, respectively.
Under setting \Rmnum{2}, CAT-Net achieves 70.88\% Dice on Abd-CT, 75.22\% Dice on Abd-MRI, and 79.36\% Dice on Card-MRI, outperforming SOTAs by 2.56\%, 2.02\% and 1.32\%, respectively. 
The consistent superiority of our method to SOTAs on three datasets and under two evaluation settings indicates the effectiveness and generalization ability of the proposed CAT-Net.
In addition, the qualitative results in Fig.~\ref{fig:vision} demonstrate that the proposed method is able to generate more accurate and detailed segmentation results compared to SOTAs.

\subsection{Ablation Study}
We conduct an ablation study to investigate the effectiveness of each component in CAT-Net.
All ablation studies are conducted on Abd-MRI under setting \Rmnum{2}.

\noindent\textbf{Effectiveness of CMAT Block:}
To demonstrate the importance of our proposed CAT-Net in narrowing the information gap between the query and supporting images and obtaining enhanced features, we conducted an ablation study.
Specifically, we compared the results of learning foreground information only from the support (\textit{S$\to$Q}) or query image (\textit{Q$\to$S}) and obtaining a single enhanced feature instead of two (\textit{S$\leftrightarrow$Q}). 
It can be observed that using the enhanced query feature (\textit{S$\to$Q}) achieves 66.72\% in Dice, outperforming only using the enhanced support feature (\textit{Q$\to$S}) by 0.74\%.
With our CMAT block, the mutual boosted support and query feature (\textit{S$\leftrightarrow$Q}) could improve the Dice by 1.90\%.
Moreover, the iteration refinement framework consistently promotes the above three variations by 0.96\%, 0.56\%, and 2.26\% in Dice, respectively.
\begin{figure}[htbp]
\begin{minipage}{.48\textwidth}
\begin{table}[H]
\centering
\setlength{\tabcolsep}{2pt}{
\begin{tabular}{c c c c c c}
\toprule
\textit{S$\to$Q} & \textit{Q$\to$S} & \textit{S$\leftrightarrow$Q} & Iter & Dice & Improve \\
\hline 
\checkmark &  &  &  & 66.72 & - \\
 & \checkmark &  &  &  65.98 & \textcolor{blue}{-0.74}\\
&  & \checkmark &  &  68.62 & \textcolor{red}{+1.90}\\
\checkmark &  &  &  \checkmark & 67.68& \textcolor{red}{+0.96}\\
&  \checkmark	& 	&\checkmark	& 66.54& \textcolor{red}{+0.56}\\ 
&  & \checkmark &	\checkmark	&\textbf{70.88}	& \textcolor{red}{+2.26}\\
\bottomrule
\vspace{0.2cm}
\end{tabular}
\caption{Effectiveness of each component. \textit{S$\to$Q} and \textit{Q$\to$S} denote one branch CAT-Net to enhance support or query feature, respectively. \textit{S$\leftrightarrow$Q} indicates applying cross attention to both \textit{S} and \textit{Q}.}
\label{tab:abl_com}}
\end{table}
\end{minipage}
\hfill
\begin{minipage}{0.48\textwidth}
 \centerline{\includegraphics[width=\textwidth]{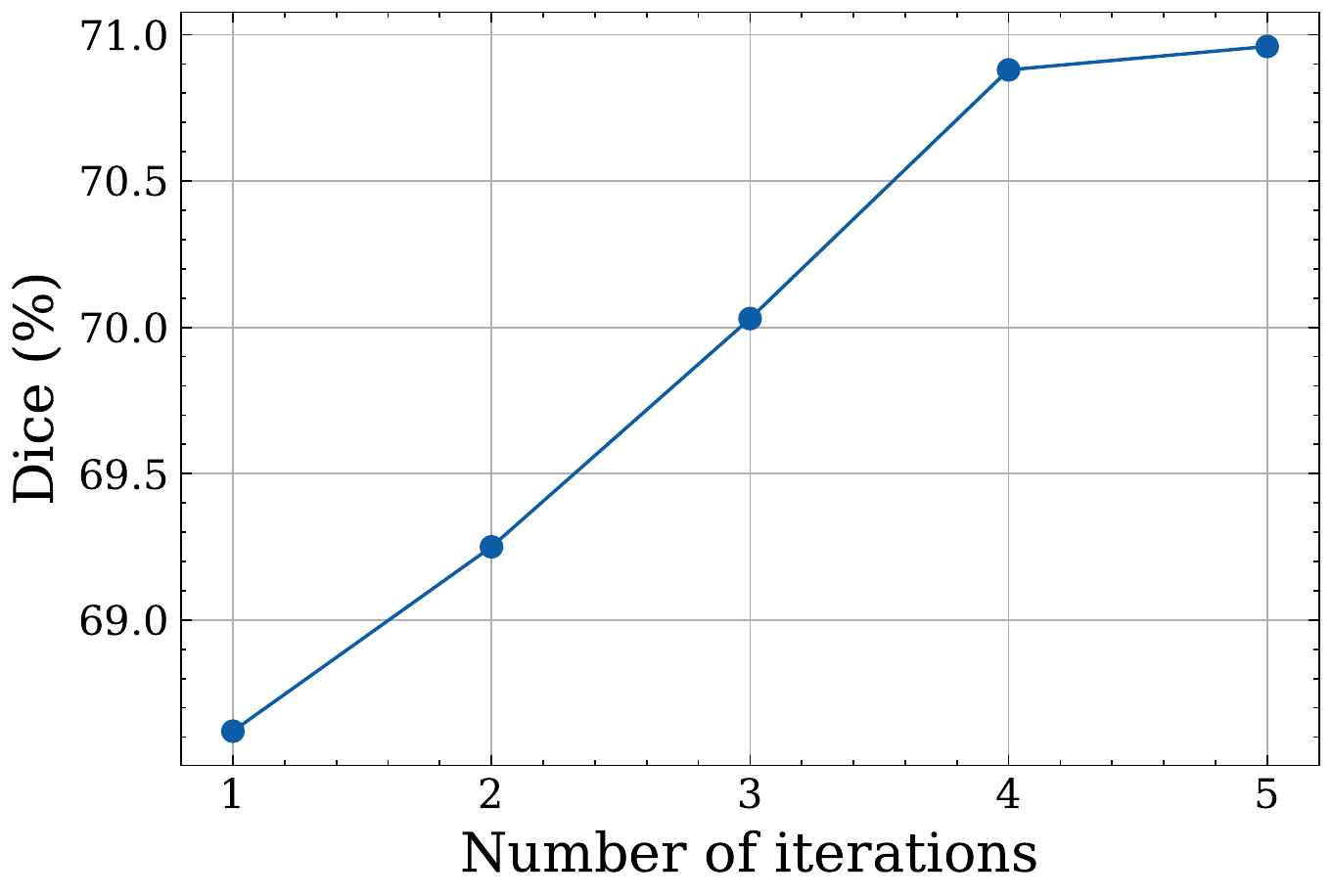}}
 \caption{The influence of different numbers of iteration CMAT modules.}
 \label{fig:abl_iter}
\end{minipage}
\end{figure}

 \noindent\textbf{Influence of Iterative Mask Refinement Block:}
To determine the optimal number of iterative refinement CMAT block, we experiment with different numbers of blocks. 
In Fig.~\ref{fig:abl_iter}, we observe that increasing the number of blocks results in improved performance, with a maximum improvement of 2.26\% in Dice when using 5 blocks.
Considering the performance gain between using 4 and 5 CMAT blocks was insignificant, we hence opt to use four CMAT blocks in our final model to strike a balance between efficiency and performance.
%------------------------------------------
% conclusion
\section{Conclusion}
In this paper, we propose CAT-Net, Cross Attention Transformer network for few-shot medical image segmentation.
Our CAT-Net enables mutual interaction between the query and support features by the cross masked attention module, enhancing the representation abilities for both of them.
Additionally, the proposed CMAT module can be iteratively applied to continually boost the segmentation performance.
Experimental results demonstrated the effectiveness of each module and the superior performance of our model to the SOTA methods.
 In the future, we plan to extend our CAT-Net from 2D to 3D networks, explore the application of our model to other medical image segmentation tasks, as well as the extension of our model to other clinical applications, such as 
 rare diseases and malformed organs, where data and annotations are scarce and costly.
%------------------------------------------
\bibliographystyle{splncs04}
\bibliography{ref.bib}
\end{document}